\documentclass{article}

 \usepackage[preprint]{neurips_2026}

\usepackage{microtype}      % microtypography
\usepackage[utf8]{inputenc} % allow utf-8 input
\usepackage[T1]{fontenc}    % use 8-bit T1 fonts
\usepackage{hyperref}       % hyperlinks
\usepackage{url}            % simple URL typesetting
\usepackage{booktabs}       % professional-quality tables
\usepackage{amsfonts}       % blackboard math symbols
\usepackage{nicefrac}       % compact symbols for 1/2, etc.
\usepackage{xcolor}         % colors
\usepackage{algorithm}
\usepackage{amsmath}
\usepackage{makecell}
\usepackage{multirow}
\usepackage{algpseudocode}
\usepackage{amssymb}
\usepackage{ragged2e}
\usepackage{graphicx}
\usepackage{svg}
\usepackage{float}

\usepackage{tikz}
\usepackage{caption}
\usepackage{amsthm}

\title{HarnessEvolve: Learning from Reference Trajectories for Reliable Agent Self-Evolution}

\author{%
  \textbf{Wen Jiang$^{*}$, Mingmin Chu$^{*}$, Yimeng Tian$^{*}$, Qianxin Zhang$^{*}$, Haofei Yang} \\
  \textbf{Rui Yang, Yang Liu, Tao Lv, Fangming Li} \\
  ICT AI Competence Center, Huawei Technologies, Shanghai, China \\
  \texttt{\{yangrui235, lifangming1\}@huawei.com} \\
  $^{*}$Equal contribution. \quad Corresponding authors: Rui Yang, Fangming Li
}

\begin{document}

\maketitle

\begin{abstract}
Self-evolving agents advance toward autonomy by optimizing their harness---prompts, skills, tools, and execution logic---based on environmental feedback. This paradigm, however, is hampered by three challenges: \textit{credit assignment failure}, where terminal success/failure feedback makes it ambiguous which step caused the error; \textit{shortcut learning}, where agents memorize task-specific patterns rather than acquire generalizable capabilities; and \textit{catastrophic forgetting}, where unguarded updates degrade previously acquired competence. In this paper, we introduce HarnessEvolve, a self-evolving framework that learns from reference trajectories to achieve reliable agent self-evolution. HarnessEvolve decouples the execution agent from the evolutionary pipeline, assigning execution, evaluation, optimization, and gating to independent agent modules, enabling generalizable and stable harness improvements. Specifically, HarnessEvolve overcomes credit assignment failure by generating reference trajectories (execution paths produced when given the ground-truth answers) and aligning failed executions against them to extract error signals, which are clustered to reveal systematic failure patterns. To prevent shortcut learning and catastrophic forgetting, candidate harness updates must pass two gates: a quality gate that filters data leakage and prompt bloat, and a performance gate that accepts each update if it improves on the current batch without degrading recent batches, with epoch-end validation on a held-out set selecting the best-performing accepted agent snapshot. We conduct extensive experiments on several benchmarks spanning open-domain and enterprise scenarios, using different models and agent frameworks. Results demonstrate that HarnessEvolve consistently outperforms state-of-the-art baselines across all benchmarks and settings, confirming reliability across task domains.
\end{abstract}

\section{Introduction}

Large language models (LLMs) have demonstrated remarkable capabilities in reasoning, planning, and tool use, enabling the development of autonomous agents that tackle complex tasks across domains ranging from software engineering to enterprise data analysis. At the core of such agents is their \emph{harness}: the prompts, skills, tools, and execution logic that collectively shape their capabilities and execution behavior.

The design of such harnesses has evolved through several waves of decreasing reliance on human priors. Early systems relied on \emph{handcrafted workflows}~\citep{wei2022cot,yao2023react}, where human experts explicitly design execution logic, search topologies, and collaboration structures, while LLMs serve as execution nodes within predefined boundaries. While this paradigm offers high controllability, it lacks flexibility: deviations from anticipated workflows can cause the entire pipeline to fail, and agents remain bounded within human-designed frameworks, unable to expand without human intervention~\citep{sumers2024cognitive}. \emph{Skill-augmented agents} relax this rigidity: humans provide skills (described in natural language) and models operate within autonomous perception-planning-action-observation loops, exemplified by systems such as Claude Code and OpenClaw. This paradigm enables dynamic trial-and-error and self-healing through iterative reasoning, but it remains bounded by the quality of human-provided skills and can become stuck in iterative retry loops when initial planning decisions are incorrect. These limitations motivate the pursuit of \emph{self-evolving agents} that can autonomously synthesize, refine, and iteratively improve their own harness from environmental feedback, progressively reducing reliance on human priors.

Despite recent progress, self-evolving agents remain difficult to deploy in real-world settings. Existing frameworks typically provide only sparse, terminal rewards (i.e., a binary success/failure signal at the end of execution), without mechanisms for fine-grained trajectory analysis to indicate which intermediate actions were correct or incorrect. Under such sparse signals, models tend to take shortcuts and overfit to specific evaluation instances, while unverified updates compound over iterations, gradually degrading prior capabilities. We identify three fundamental challenges that hinder reliable self-evolution:

\begin{itemize}
    \item \textbf{Credit Assignment Failure:} In long-horizon tasks, agents receive only binary success/failure signals at the end. When a failure occurs, the agent cannot identify which specific action caused it, because later errors are downstream consequences of earlier mistakes. Solely reflecting on the whole trajectory may misidentify the root cause, leaving optimization directionless.
    \item \textbf{Shortcut Learning:} Without safeguards, agents tend to hardcode answers or inject excessive task-specific in-context examples into the harness, causing data leakage and prompt bloat that inflate training accuracy without genuine capability gain.
    \item \textbf{Catastrophic Forgetting:} Since each harness update is optimized on a subset of trajectory data, it may conflict with previously acquired competence, gradually degrading existing capabilities over multiple iterations.
\end{itemize}

These challenges cannot be resolved by designing yet another iterative prompt optimization algorithm; they necessitate a system-level architectural shift. To this end, we introduce HarnessEvolve, a self-evolving framework that learns from reference trajectories to achieve reliable agent self-evolution. Rather than entangling execution and optimization in a single reasoning loop, HarnessEvolve decouples the execution agent from the evolutionary pipeline, assigning execution, evaluation, optimization, and gating to independent modules, ensuring that optimization decisions are auditable and cannot be gamed by the execution agent.

Concretely, the execution agent runs tasks using its harness, which serves as the target of optimization. An evaluation agent assesses both the accuracy of the execution agent and the reliability of reference trajectories, the latter produced by the execution agent given the question and its ground-truth answer. Based on these assessments, an optimization agent analyzes failed executions by comparing them against reference trajectories to identify the first point of divergence, isolating the root-cause action from downstream cascading errors; these error signals are aggregated via error clustering to identify systematic failure patterns. A gate agent then inspects candidate updates for data leakage (by comparing content against failed training instances) and prompt bloat (by counting injected in-context examples), rejecting updates that exhibit either. Finally, the performance gate accepts each candidate only if it improves on the current batch without degrading recent batches, with epoch-end validation on a held-out set selecting the best-performing accepted agent snapshot.

We evaluate HarnessEvolve against several self-evolving baselines (GEPA~\citep{gepa2025}, ACE~\citep{zhang2026agentic}, and SkillOpt~\citep{skillopt2026}), each optimizing only a single component of the harness (e.g., SkillOpt optimizes solely \texttt{skill.md}), on five benchmarks spanning open-domain tasks (SearchQA~\citep{dunn2017searchqa}, OfficeQA~\citep{opsahl2026officeqa}, SpreadsheetBench~\citep{ma2024spreadsheetbench}) and specialized enterprise QA and text-to-SQL tasks (CloudCoreNetwork-QA and Wireless-QA, two in-house datasets from cloud core network and wireless network domains). HarnessEvolve consistently outperforms all baselines on all five benchmarks in terms of accuracy. Notably, on CloudCoreNetwork-QA, HarnessEvolve outperforms the strongest baseline by 21.6 percentage points, demonstrating substantial gains on complex multi-skill enterprise tasks. In summary, the main contributions of HarnessEvolve are fourfold:
\begin{itemize}
    \item \textbf{Reference-Guided Error Diagnosis:} A diagnostic method that overcomes credit assignment failure by comparing failed executions against reference trajectories. The resulting error signals are aggregated via error clustering to identify systematic failure patterns.
    \item \textbf{Quality Gate against Shortcut Learning:} A filtering mechanism that inspects candidate updates for data leakage and prompt bloat. Only harness updates that pass both checks are retained for downstream validation.
    \item \textbf{Performance Gate for Stable Evolution:} A validation-gated update protocol that mitigates catastrophic forgetting. Each harness candidate is accepted into an agent snapshot pool only if it improves on the current batch without degrading recent batches, with epoch-end validation on a held-out set selecting the best-performing snapshot.
    \item \textbf{Comprehensive and Reliable Harness Self-Evolution:} Built on a decoupled multi-agent architecture, HarnessEvolve constitutes a comprehensive and reliable harness self-evolution framework that optimizes the entire harness rather than a single component, achieving state-of-the-art accuracy on both in-house enterprise and open-source benchmarks.
\end{itemize}

\section{Related Work}
Handcrafted agents rely on human-designed static workflows, where experts define prompt templates, reasoning structures, search topologies, and collaborative pipelines~\citep{wei2022cot,yao2023react,press2023selfask,yao2024tot,besta2024got,hong2024metagpt,qian2024chatdev,wu2023autogen,schick2023toolformer}. Representative systems include LangChain/LangGraph and visual builders like Flowise and Dify. While offering high controllability, this paradigm confines agents within predefined boundaries and scales poorly as task complexity grows~\citep{khattab2024dspy,zhou2024webarena,sumers2024cognitive,liu2024agentbench,ma2024agentboard}.

Skill-augmented agents relax this rigidity: humans provide skills in natural language while models operate within autonomous reasoning loops~\citep{yao2023react}, exemplified by Claude Code and OpenClaw. While enabling dynamic trial-and-error, this paradigm faces a critical limitation: agent performance remains heavily dependent on the quality of human-provided skills~\citep{skillopt2026,liang2026skillnet}, creating a bottleneck for scalability. These limitations motivate self-evolving systems that can autonomously synthesize, refine, and iteratively improve their own harness from environmental feedback, progressively reducing reliance on human priors.

Self-evolving agents aim to autonomously optimize their inner logic. Current approaches optimize along multiple dimensions: workflow-level architecture design~\citep{zhang2025aflow,liu2025sew}, prompt-level optimization (e.g., OPRO~\citep{yang2023opro}, APO~\citep{pryzant2023protegi}, and GEPA~\citep{gepa2025}), context-level evolution such as Reflexion~\citep{shinn2023reflexion}, ExpeL~\citep{zhao2024expel}, and ACE~\citep{zhang2026agentic}, and tool/skill-level refinement including Voyager~\citep{wang2023voyager} and SkillOpt~\citep{skillopt2026}. However, each optimizes only a single component of the harness independently, leaving the rest untouched and limiting the space of achievable improvements.

More recently, methods that directly evolve a more comprehensive harness system have emerged: Self-Harness~\citep{zhang2026selfharness} introduces an iterative loop of weakness mining, harness proposal, and regression-testing-based validation, while Agentic Harness Engineering (AHE)~\citep{lin2026ahe} evolves coding-agent harnesses through observability-driven falsifiable contracts. Despite their advances, these methods keep credit assignment unresolved (i.e., no reference-based root-cause localization), leave shortcut learning unguarded (i.e., no inspection of data leakage or prompt bloat), and only partially mitigate catastrophic forgetting (i.e., no agent snapshot pool)---gaps that HarnessEvolve addresses.

\section{Methodology}
\label{sec:methodology}
\subsection{Overview}
\label{sec:overview}

We consider the problem of agent self-evolution. Given a task corpus $\mathcal{T} = \{(q_i, a_i^*)\}_{i=1}^{M}$ of $M$ query-answer pairs (where $a_i^*$ is the ground-truth answer to query $q_i$) split into training $\mathcal{T}_{\mathrm{train}}$, validation $\mathcal{T}_{\mathrm{val}}$, and test $\mathcal{T}_{\mathrm{test}}$ sets, and a base execution agent $\mathcal{A}_{\mathrm{exec}}$ powered by a frozen LLM that runs tasks from $\mathcal{T}$, the goal is to find the optimal execution agent $\mathcal{A}^*$ by iteratively optimizing its harness---prompts, skills, tools, and execution logic---on $\mathcal{T}_{\mathrm{train}}$ without human intervention, such that accuracy on the validation set $\mathcal{T}_{\mathrm{val}}$ is maximized. The test set $\mathcal{T}_{\mathrm{test}}$ is used to evaluate the final accuracy of $\mathcal{A}^*$.

HarnessEvolve addresses this through an architecture (Figure~\ref{fig:framework}) that separates the execution agent from the evolutionary pipeline, assigning execution, evaluation, optimization, and gating to independent agent modules:

\begin{figure}[t]
    \centering
    \includegraphics[width=0.9\columnwidth]{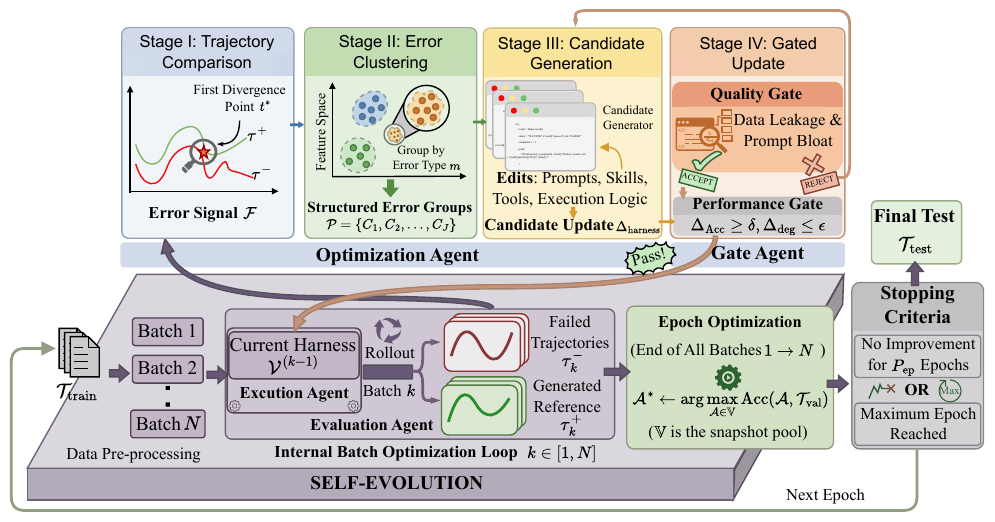}
    \caption{Overview of the HarnessEvolve self-evolution framework. Four agents---$\mathcal{A}_{\mathrm{exec}}$, $\mathcal{A}_{\mathrm{eval}}$, $\mathcal{A}_{\mathrm{opt}}$, $\mathcal{A}_{\mathrm{gate}}$---collaborate in a batch-driven loop: the execution agent produces trajectories, the evaluation agent identifies failures and verifies references, the optimization agent diagnoses and clusters errors to generate candidate harness updates, and the gate agent enforces quality and performance checks before accepting updates into the snapshot pool $\mathbb{V}$.}
    \label{fig:framework}
\end{figure}

\begin{itemize}
    \item \textbf{Execution:} The execution agent $\mathcal{A}_{\mathrm{exec}}$, the target of optimization, runs tasks from $\mathcal{T}_{\mathrm{train}}$ using its harness. For each task in $\mathcal{T}_{\mathrm{train}}$, the execution agent is also run with the question and its ground-truth answer to produce reference trajectories that reach the correct answer.

    \item \textbf{Evaluation:} The evaluation agent $\mathcal{A}_{\mathrm{eval}}$ performs two assessments: (1) it evaluates the task accuracy of the execution agent and identifies failed trajectories; (2) it verifies that reference trajectories are genuine execution paths rather than shortcuts where the agent directly returns the known answer, ensuring reference trajectory reliability.

    \item \textbf{Optimization:} The optimization agent $\mathcal{A}_{\mathrm{opt}}$ analyzes failed trajectories and, when reference trajectories are available, compares them to identify the root cause of failures. It clusters batch-level errors to identify systematic error patterns, and based on these diagnoses, proposes targeted modifications for the harness of $\mathcal{A}_{\mathrm{exec}}$.

    \item \textbf{Gating:} The gate agent $\mathcal{A}_{\mathrm{gate}}$ enforces two checks for the harness modifications proposed by $\mathcal{A}_{\mathrm{opt}}$: (1) a data leakage and prompt bloat inspection, where modifications violating these constraints are returned to $\mathcal{A}_{\mathrm{opt}}$ for revision; (2) a performance check, where a modification is accepted only if it improves accuracy on the current batch without degrading performance on $R$ previously executed batches.
\end{itemize}

To optimize the harness of the base execution agent $\mathcal{A}_{\mathrm{exec}}$, HarnessEvolve runs in a batch-driven loop. At the start of each epoch, $\mathcal{T}_{\mathrm{train}}$ is randomly permuted and partitioned into mini-batches $\mathcal{B} = \mathrm{PermuteAndPartition}(\mathcal{T}_{\mathrm{train}}, b) = \{B_1, B_2, \dots\}$, where $b$ is the batch size and $B_j$ denotes the $j$-th mini-batch. For each mini-batch $B_j$, the four agents collaborate in sequence: the execution agent produces trajectories, the evaluation agent identifies failed trajectories, the optimization agent generates candidate modifications, and the gate agent validates and accepts beneficial modifications, storing each accepted harness as a snapshot in a pool $\mathbb{V}$. After all mini-batches in $\mathcal{B}$ complete, the best-performing snapshot on $\mathcal{T}_{\mathrm{val}}$ is selected as the final harness of the epoch, and the optimization loop continues for several epochs.

\subsection{Execution}
\label{sec:execution}

The execution agent $\mathcal{A}_{\mathrm{exec}}$ runs tasks from $\mathcal{T}_{\mathrm{train}}$ using its harness, the collection of prompts, skills, tools, and execution logic that defines its behavioral space. For each query $q_i$ in $\mathcal{T}_{\mathrm{train}}$, $\mathcal{A}_{\mathrm{exec}}$ produces a trajectory $\tau_i = [(o_1, a_1), (o_2, a_2), \dots, (o_{T_i}, a_{T_i})]$, where $o_t$ and $a_t$ denote the observation and action at step $t$, and $T_i$ is the trajectory length.

To address credit assignment failure, prior to the optimization loop, $\mathcal{A}_{\mathrm{exec}}$ is also run on all tasks in $\mathcal{T}_{\mathrm{train}}$ with both the questions and their ground-truth answers. For each task, $\mathcal{A}_{\mathrm{exec}}$ attempts up to $T_{\mathrm{att}}$ times; after each attempt, the evaluation agent $\mathcal{A}_{\mathrm{eval}}$ verifies whether the produced trajectory is genuine (Section~\ref{sec:evaluation}). Once a trajectory passes verification, it is stored as the verified reference trajectory $\tau^+_i$ in a global cache $\mathcal{C}$. If all $T_{\mathrm{att}}$ attempts fail to produce the correct answer or to pass $\mathcal{A}_{\mathrm{eval}}$'s verification, the task is left without a reference trajectory and falls back to single-trajectory analysis during optimization. These reference trajectories provide successful execution paths that can be compared against failed trajectories to pinpoint the first point of divergence (the specific action where $\mathcal{A}_{\mathrm{exec}}$ went wrong) rather than attributing failure to the entire trajectory. This reference-guided comparison enables fine-grained error diagnosis that cannot be achieved by reflecting on failed trajectories alone.

\subsection{Evaluation}
\label{sec:evaluation}

The evaluation agent $\mathcal{A}_{\mathrm{eval}}$ serves a dual role: assessing execution accuracy during optimization and verifying reference trajectory genuineness during reference trajectory generation. Both assessments are essential for reliable error diagnosis.

\textbf{Accuracy Assessment.} During each optimization batch, $\mathcal{A}_{\mathrm{eval}}$ collects the trajectories produced by $\mathcal{A}_{\mathrm{exec}}$ on the current mini-batch $B_j$, and compares the final output of each trajectory against the corresponding ground-truth answer $a_i^*$. Trajectories that fail to produce the correct answer are collected as the failed set $F_j = \{(q_i, \tau^-_i)\}$, which serves as input to the error diagnosis pipeline (Section~\ref{sec:optimization}). Moreover, the accuracy assessment on $B_j$ also determines whether the candidate harness updates are accepted by the performance gate (Section~\ref{sec:gating}).

\textbf{Reference Verification.} Since $\mathcal{A}_{\mathrm{exec}}$ is given the ground-truth answer during reference trajectory generation, it might produce a trajectory that shortcuts through the reasoning: for example, directly outputting the answer without invoking the necessary tools or following the expected workflow. To prevent this, $\mathcal{A}_{\mathrm{eval}}$ verifies each produced trajectory before it enters the cache $\mathcal{C}$. The verification checks that the trajectory follows a legitimate reasoning chain: $\mathcal{A}_{\mathrm{exec}}$ invokes appropriate tools, processes intermediate observations, and arrives at the answer through valid steps rather than trivially restating the provided answer. Only verified trajectories are stored as $\tau^+_i$; unverified ones are rejected, and $\mathcal{A}_{\mathrm{exec}}$ continues to the next attempt (Section~\ref{sec:execution}).

\subsection{Optimization}
\label{sec:optimization}

The optimization agent $\mathcal{A}_{\mathrm{opt}}$ analyzes failed trajectories, compares them against reference trajectories to identify root causes, clusters errors into systematic patterns, and generates candidate modifications to the harness.

\textbf{Trajectory Comparison.} Given a failed trajectory $\tau^-_i$, the question $q_i$, and a verified reference trajectory $\tau^+_i$ (if available), $\mathcal{A}_{\mathrm{opt}}$ produces a structured error signal $\mathcal{F}_i = (s_i, m_i, h_i)$, where $s_i$ is the severity, $m_i$ is the error cause (e.g., tool hallucination, argument omission, premature termination), and $h_i$ is a natural-language fix hint. When $\tau^+_i$ is available, $\mathcal{A}_{\mathrm{opt}}$ can identify more fine-grained differences by comparing $\tau^+_i$ and $\tau^-_i$, such as the first action divergence point $t_{i}^{*}$ (the earliest step at which the action in $\tau^-_i$ deviates from that in $\tau^+_i$), enabling more accurate error cause identification. When no reference trajectory is available, $\mathcal{A}_{\mathrm{opt}}$ analyzes only $\tau^-_i$ and $q_i$, outputting its best assessment of $s_i$, $m_i$, and $h_i$ without divergence-point localization. While less precise than reference-guided diagnosis, this fallback ensures that all failed instances contribute to the optimization signal.

\textbf{Error Clustering.} Individual error signals $\mathcal{F}_i$ are instance-level diagnoses that, if processed independently, would lead to fragmented and potentially conflicting modifications. To address this, $\mathcal{A}_{\mathrm{opt}}$ clusters error signals $\{\mathcal{F}_i\}_{i=1}^{|F_j|}$ by error cause $m_i$ into groups $\mathcal{P} = \{C_1, \dots, C_K\}$, where each $C_k = (\bar{s}_k, \bar{m}_k, \bar{e}_k, \bar{h}_k)$: $\bar{s}_k$ is the aggregated severity level, $\bar{m}_k$ is the shared error cause, $\bar{e}_k$ contains representative failed trajectories for $\bar{m}_k$, and $\bar{h}_k$ is the suggested fix direction derived from the fix hints of its members. This aggregation enables the optimization agent to identify systematic failure patterns and generate coherent modifications. The clustering enforces three principles:
\begin{itemize}
    \item \textbf{Cause-Based Grouping:} Groups are formed by error cause $m_i$ rather than coarse severity labels, preventing semantically distinct errors from being merged. Thus, each cluster corresponds to a coherent fix strategy, enabling $\mathcal{A}_{\mathrm{opt}}$ to generate targeted modifications that uniformly address all instances within the cluster.
    \item \textbf{Root-Cause Priority:} When a reference trajectory is available, $\mathcal{A}_{\mathrm{opt}}$ is instructed to prioritize the first action divergence point $t_{i}^{*}$ between $\tau^+_i$ and $\tau^-_i$ as the root error cause, since subsequent divergence points are likely cascading effects of the initial error rather than independent root causes. Clustering by the root cause ensures that $\mathcal{A}_{\mathrm{opt}}$ addresses the original failure rather than its downstream symptoms.
    \item \textbf{Long-Tail Protection:} Unlike conventional distance-based clustering (e.g., K-Means)~\citep{xiong2006kmeans}, which forces every sample into a fixed number of clusters, $\mathcal{A}_{\mathrm{opt}}$ preserves single-member clusters, ensuring rare but critical failure patterns are not absorbed into dominant clusters~\citep{liu2019large,feldman2020does}.
\end{itemize}

The complete diagnosis and clustering procedure is formalized in Algorithm~\ref{alg:unified_pipeline_clean} (Appendix).

\textbf{Candidate Generation.} The clustering result $\mathcal{P}$ is provided to $\mathcal{A}_{\mathrm{opt}}$, which generates targeted modifications to the current harness. $\mathcal{A}_{\mathrm{opt}}$ is instructed to prioritize root causes with more instances (i.e., clusters containing more error signals $\mathcal{F}_i$) and higher severity. Based on the root cause, severity, and fix direction of each cluster, the agent synthesizes specific edits to the harness of $\mathcal{A}_{\mathrm{exec}}$ to address the root cause. The output is a candidate harness update set $\Delta_{\mathrm{harness}}$, which is passed to the gate agent for validation.

\subsection{Gating}
\label{sec:gating}

\textbf{Quality Gate.} The gate agent $\mathcal{A}_{\mathrm{gate}}$ ensures generalization by verifying that the optimization agent has not introduced shortcuts into the harness. The gate agent inspects candidate updates for data leakage and prompt bloat. For data leakage, the gate agent checks whether the edits proposed by the optimization agent directly embed failed queries and their ground-truth answers into the harness files, using LLM-as-judge with a score ranging from 0 to 1; if the score exceeds the threshold $\eta_{\mathrm{leak}}$, the update is rejected. For prompt bloat, the gate agent counts newly injected in-context examples; if the count exceeds the threshold $\eta_{\mathrm{blo}}$, the update is rejected. A candidate update passes the quality gate if and only if, for every file $f$ in $\Delta_{\mathrm{harness}}$ (the set of harness files edited by the optimization agent), the data leakage score of $f$ does not exceed $\eta_{\mathrm{leak}}$, and the number of injected in-context examples in $f$ does not exceed $\eta_{\mathrm{blo}}$. An update that passes yields a verified update $\Delta_{\mathrm{verified}}$; a rejected update is returned to the optimization agent along with a rejection reason $\rho$ (a natural-language explanation of the violated constraint) for revision. If the gate remains unsatisfied after a fixed number of revisions, the update is abandoned (i.e., $\Delta_{\mathrm{verified}} = \emptyset$). The gate agent operates as an isolated judge, structurally decoupled from the optimization agent, preventing self-serving evaluation.

\textbf{Performance Gate.} The verified update $\Delta_{\mathrm{verified}}$ is used to construct a candidate agent $\mathcal{A}_{\mathrm{candidate}} = \mathrm{Build}(\mathcal{A}_{\mathrm{current}}, \Delta_{\mathrm{verified}})$, where $\mathcal{A}_{\mathrm{current}}$ denotes the current version of $\mathcal{A}_{\mathrm{exec}}$ being optimized. The candidate is evaluated on the current mini-batch $B_j$ and a buffer of $R$ recent mini-batches $\{B_{\max(1,\, j-R)}, \dots, B_{j-1}\}$ (empty when $j=1$). A candidate is accepted if and only if it improves accuracy on $B_j$ by at least a margin $\delta$ while not degrading on any recent batch beyond tolerance $\epsilon$:
\begin{equation}
\label{eq:update_decision}
\mathcal{A}_{\mathrm{candidate}} \text{ is accepted} \iff
\begin{cases}
\mathrm{Acc}(\mathcal{A}_{\mathrm{candidate}}, B_j) - \mathrm{Acc}(\mathcal{A}_{\mathrm{current}}, B_j) \geq \delta, \\
\displaystyle\max_{l \in [\max(1,\, j-R),\, j-1]} \big[\mathrm{Acc}(\mathcal{A}_{\mathrm{current}}, B_l) - \mathrm{Acc}(\mathcal{A}_{\mathrm{candidate}}, B_l)\big] \leq \epsilon.
\end{cases}
\end{equation}
Accepted candidates are stored as snapshots in the pool $\mathbb{V}$ and immediately become the active agent $\mathcal{A}_{\mathrm{current}}$ for subsequent batches; performance-gate rejections increment a consecutive-rejection counter $c$. If $c$ reaches the patience threshold $P_{\mathrm{batch}}$, the current epoch terminates early (bypassing remaining batches) and proceeds directly to epoch-end selection, thereby mitigating redundant computation upon convergence.

\textbf{Epoch-End Selection.} At epoch completion, all snapshots accumulated in the pool $\mathbb{V}$ are evaluated on the validation set $\mathcal{T}_{\mathrm{val}}$, and the best-performing one is selected:
\begin{equation}
\begin{aligned}
\mathcal{A}_{\mathrm{current}} &\leftarrow \arg\max_{\mathcal{A} \in \mathbb{V}} \mathrm{Acc}(\mathcal{A}, \mathcal{T}_{\mathrm{val}}), \\
\mathcal{A}^* &\leftarrow \mathcal{A}_{\mathrm{current}}.
\end{aligned}
\end{equation}
This two-tiered architecture (batch-level updates within each epoch and epoch-level selection on $\mathcal{T}_{\mathrm{val}}$) ensures that the final agent is no worse than the best-performing snapshot in the pool. The optimization loop terminates when the epoch count reaches $E$, or when the validation accuracy of $\mathcal{A}^*$ fails to improve for $P_{\mathrm{ep}}$ consecutive epochs, indicating convergence.

The complete HarnessEvolve self-evolution procedure (encompassing execution, evaluation, optimization, and gating) is formalized in Algorithm~\ref{alg:evolution_loop_clean} (Appendix). After all epochs complete, the final agent $\mathcal{A}^*$ is evaluated on the test set $\mathcal{T}_{\mathrm{test}}$ to report accuracy.

\section{Experiments}

\begin{table}[t]
\centering
%\small
\caption{Main results on in-house datasets. Accuracy (\%) is reported. Best results are in \textbf{bold}.}
\label{tab:internal}
\setlength{\tabcolsep}{8pt}
\begin{tabular}{lcccc}
\toprule
\multirow{2}{*}{\textbf{Method}} & \multicolumn{2}{c}{\textbf{CloudCoreNetwork-QA}} & \multicolumn{2}{c}{\textbf{Wireless-QA}} \\
\cmidrule(lr){2-3} \cmidrule(lr){4-5}
& \textbf{Qwen} & \textbf{DeepSeek} & \textbf{Qwen} & \textbf{DeepSeek} \\
\midrule
Base     & 43.4          & 47.5 & 79.0 & 85.9 \\
GEPA     & 65.3          & 57.6 & 82.4 & 86.5 \\
ACE      & 59.3          & 64.6 & 84.3 & 90.1 \\
SkillOpt & 61.9          & 65.3 & 89.0 & 89.3 \\
HarnessEvolve & \textbf{86.9} & \textbf{85.9} & \textbf{89.7} & \textbf{92.8} \\
\bottomrule
\end{tabular}
\end{table}

\subsection{Experimental Setup}
\textbf{Datasets.} We evaluate all methods on three open-source datasets (SearchQA~\citep{dunn2017searchqa}, OfficeQA~\citep{opsahl2026officeqa}, SpreadsheetBench~\citep{ma2024spreadsheetbench}) and two in-house datasets (CloudCoreNetwork-QA and Wireless-QA) to assess generalization across open-domain and enterprise scenarios. SearchQA is an open-domain QA dataset where agents retrieve and synthesize information from a search engine. OfficeQA is an enterprise-level document reasoning benchmark where agents parse, retrieve, and reason across unstructured text and tabular data to answer complex questions. SpreadsheetBench is a spreadsheet manipulation benchmark where agents perform data and formula operations. CloudCoreNetwork-QA and Wireless-QA are constructed from cloud core network and wireless network domains, where agents perform network-domain QA and text-to-SQL tasks such as KPI metric querying and network element configuration querying. Queries in both datasets require selecting and combining multiple skills.

\textbf{Baselines.} We compare HarnessEvolve against three state-of-the-art agent self-optimization frameworks, each operating at a different granularity: GEPA~\citep{gepa2025} performs prompt-level optimization via reflective prompt evolution, ACE~\citep{zhang2026agentic} optimizes at the context level, and SkillOpt~\citep{skillopt2026} refines skill descriptions at the text level. All baselines are deployed under identical environmental constraints and an initial harness to ensure fair comparison. We also report the accuracy of the agent with the unoptimized harness (Base) as a reference.

\textbf{Implementation Details.} We use two models of different scales as reasoning engines: Qwen3.6-27B, post-trained with domain-specific fine-tuning, and DeepSeek-V4-Flash, used without fine-tuning. On the in-house datasets (CloudCoreNetwork-QA and Wireless-QA), both models are used under our in-house LAMAgent framework, where the harness being optimized includes the full project code, skills, prompts, and tools. On the open-source datasets (SearchQA, OfficeQA, and SpreadsheetBench), DeepSeek-V4-Flash is used under OpenClaw, where the harness being optimized comprises skills, \texttt{AGENTS.md}, \texttt{SOUL.md}, and tools. Unlike SkillOpt, which optimizes only \texttt{skill.md}, HarnessEvolve optimizes the entire skill directory including reference files and scripts. All methods use the same training, validation, and test split on each dataset. We set max reference attempts $T_{\mathrm{att}}=5$ in reference trajectory generation (Section~\ref{sec:execution}), and batch patience $P_{\mathrm{batch}}=10$, epoch patience $P_{\mathrm{ep}}=5$, epoch limit $E=20$, batch size $b=40$, replay buffer size $R=2$, step margin $\delta=0.0$, degradation tolerance $\epsilon=0.025$, data leakage threshold $\eta_{\mathrm{leak}}=0.8$, prompt bloat threshold $\eta_{\mathrm{blo}}=5$, and revision limit $T_{\mathrm{rev}}=3$ in Algorithm~\ref{alg:evolution_loop_clean} (Appendix).

\subsection{Results and Analysis}

\begin{table}[t]
\centering
%\small
\caption{Main results on open-source datasets. Accuracy (\%) is reported. Best results are in \textbf{bold}.}
\label{tab:external}
\setlength{\tabcolsep}{8pt}
\begin{tabular}{lccc}
\toprule
\textbf{Method} & \textbf{SearchQA} & \textbf{OfficeQA} & \textbf{SpreadsheetBench} \\
\midrule
Base & 86.5 & 62.8 & 44.3 \\
GEPA & 88.6 & 64.0 & 69.6 \\
ACE & 90.0 & 68.9 & 52.2 \\
SkillOpt & 89.4 & 66.9 & 74.6 \\
HarnessEvolve & \textbf{92.9} & \textbf{70.9} & \textbf{76.4} \\
\bottomrule
\end{tabular}
\end{table}

\begin{table}[t]
\centering
%\small
\caption{Cross-framework transfer: skills optimized on OpenClaw, evaluated on four frameworks. Accuracy (\%) is reported. Best results are in \textbf{bold}.}
\label{tab:transfer}
\setlength{\tabcolsep}{5pt}
\begin{tabular}{lcccccc}
\toprule
\multirow{2}{*}{\textbf{Framework}} & \multicolumn{2}{c}{\textbf{SearchQA}} & \multicolumn{2}{c}{\textbf{OfficeQA}} & \multicolumn{2}{c}{\textbf{SpreadsheetBench}} \\
\cmidrule(lr){2-3} \cmidrule(lr){4-5} \cmidrule(lr){6-7}
 & Base & HarnessEvolve & Base & HarnessEvolve & Base & HarnessEvolve \\
\midrule
Hermes & \textbf{95.0} & \textbf{95.0} & 77.9 & \textbf{80.8} & 72.1 & \textbf{73.2} \\
OpenCode & 90.0 & \textbf{92.7} & 75.6 & \textbf{80.3} & 57.5 & \textbf{87.9} \\
LAMAgent & 92.9 & \textbf{94.3} & 74.4 & \textbf{75.0} & 45.0 & \textbf{80.0} \\
DeepSeek Harness & 93.6 & \textbf{95.0} & 79.1 & \textbf{79.7} & 56.4 & \textbf{86.8} \\
\bottomrule
\end{tabular}
\end{table}

\textbf{Main Results.}
Tables~\ref{tab:internal} and~\ref{tab:external} compare HarnessEvolve against three baselines on the in-house and open-source datasets, respectively. HarnessEvolve consistently achieves the highest accuracy across all evaluated settings. Notably, on the in-house datasets, HarnessEvolve improves accuracy on CloudCoreNetwork-QA with Qwen3.6-27B from 43.4\% (Base) to 86.9\%, outperforming the strongest baseline GEPA at 65.3\% by 21.6 percentage points, and reaches 92.8\% on Wireless-QA with DeepSeek-V4-Flash, surpassing the best baseline ACE at 90.1\% by 2.7 percentage points. On the open-source datasets, HarnessEvolve consistently achieves significant accuracy improvements over all baselines across all three benchmarks. These improvements on both enterprise and open-domain benchmarks confirm that the gains from reference-guided diagnosis and comprehensive harness optimization generalize across task domains and model settings.

\begin{figure}[t]
\centering
\includegraphics[width=0.87\columnwidth]{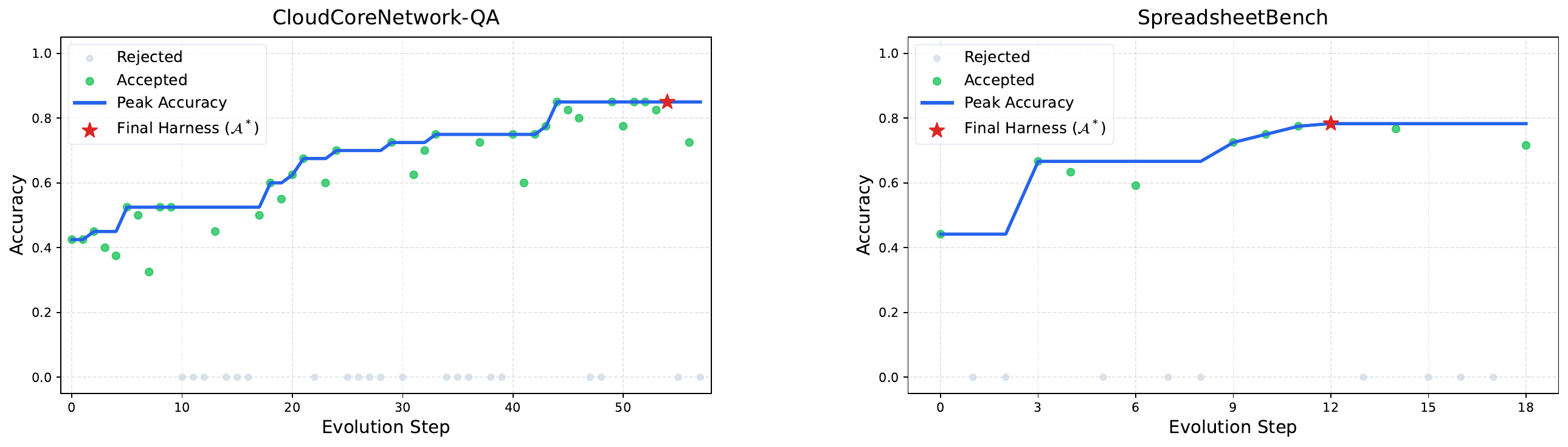}
\caption{HarnessEvolve self-evolution curves on CloudCoreNetwork-QA and SpreadsheetBench. Green points denote accepted snapshots that pass both the quality gate and performance gate and are saved to the snapshot pool $\mathbb{V}$; gray points denote rejected candidates. The blue curve tracks the peak accuracy in $\mathbb{V}$. The red star marks the final harness $\mathcal{A}^*$.}
\label{fig:evolution}
\end{figure}

\begin{table}[t]
\centering
%\small
\caption{Ablation study results on CloudCoreNetwork-QA (Qwen3.6-27B). Accuracy (\%) is reported. Best results are in \textbf{bold}.}
\label{tab:ablation}
\setlength{\tabcolsep}{8pt}
\begin{tabular}{lc}
\toprule
\textbf{Variant} & \textbf{CloudCoreNetwork-QA} \\
\midrule
HarnessEvolve (full) & \textbf{86.9} \\
M1: w/o reference trajectory & 57.8 \\
M2: w/o error clustering & 68.6 \\
M3: w/o quality gate & 80.1 \\
\bottomrule
\end{tabular}
\end{table}

\textbf{Cross-Framework Generalization.}
To evaluate whether the harness optimized by HarnessEvolve can transfer across frameworks, we take the skills optimized on SearchQA, OfficeQA, and SpreadsheetBench and deploy them directly on four frameworks to test on $\mathcal{T}_{\mathrm{test}}$, namely Hermes, OpenCode, LAMAgent, and DeepSeek Harness, without re-optimization. As shown in Table~\ref{tab:transfer}, HarnessEvolve-optimized skills consistently improve or maintain accuracy over Base across all four frameworks. These results indicate that HarnessEvolve produces transferable improvements rooted in general error patterns rather than framework-specific shortcuts.

\textbf{Self-Evolution Curve.}\label{sec:evolution}
Figure~\ref{fig:evolution} traces the self-evolution procedures of HarnessEvolve, with each point corresponding to one optimization step evaluated on $\mathcal{T}_{\mathrm{val}}$. A substantial fraction of candidate updates are rejected by the two-tier gating mechanism, indicating effective filtering of low-quality modifications. The peak accuracy in the snapshot pool $\mathbb{V}$ increases gradually throughout evolution, reaching its highest point at the final harness $\mathcal{A}^*$. This also demonstrates that the gating mechanism ensures strict quality control while retaining only beneficial updates.

\textbf{Ablation Study.}\label{sec:ablation}
To assess the contribution of each component in HarnessEvolve, we conduct ablation studies on CloudCoreNetwork-QA using Qwen3.6-27B. We consider three variants:
\begin{itemize}
    \item \textbf{M1} (w/o reference trajectory): bypassing reference trajectory comparison and performing error diagnosis solely on failed trajectories. This variant isolates the contribution of reference-guided error diagnosis, reverting to a coarser single-trajectory diagnosis.
    \item \textbf{M2} (w/o error clustering): optimizing directly over instance-level error signals $\mathcal{F}_i$ without grouping. This variant tests whether aggregating errors into systematic patterns is necessary, as processing each error signal independently may lead to fragmented and conflicting modifications.
    \item \textbf{M3} (w/o quality gate): removing the quality gate, accepting all candidate updates without filtering for data leakage and prompt bloat. This variant evaluates whether the quality gate meaningfully prevents shortcut learning, or whether the optimization agent alone can produce generalizable updates.
\end{itemize}

Table~\ref{tab:ablation} presents the ablation results. M1, which removes reference trajectory comparison, yields the largest degradation from 86.9\% to 57.8\%, confirming that reference-guided diagnosis is the most critical component. Without this comparison, the agent misidentifies root causes and generates ineffective modifications. M2, which removes error clustering, shows significant degradation to 68.6\%, confirming that aggregating instance-level errors into systematic patterns is essential for coherent modifications. M3, which removes the quality gate, shows moderate degradation to 80.1\%, indicating that the quality gate helps prevent shortcut learning and overfitting to training data.

\textbf{Qualitative Analysis.}\label{sec:qualitative}
To inspect what HarnessEvolve modifies during self-evolution, Figure~\ref{fig:optimization} (Appendix) compares the harness of the execution agent before and after optimization. The optimized harness exhibits targeted edits that span the entire execution agent, covering \texttt{skill.md} files, Python scripts within skills, prompt instructions, tool-argument specifications, and code related to agent execution logic, instead of being confined to a single component. This contrasts with SkillOpt~\citep{skillopt2026}, which optimizes only \texttt{skill.md}, and corroborates the comprehensiveness of the harness optimization of HarnessEvolve.

\section{Conclusion}

We have presented HarnessEvolve, a self-evolving framework that learns from reference trajectories for reliable agent self-evolution. By decoupling execution, evaluation, optimization, and gating into independent modules, HarnessEvolve addresses three fundamental challenges: credit assignment failure via reference-guided error diagnosis, shortcut learning via the quality gate, and catastrophic forgetting via the performance gate with epoch-end validation. Extensive evaluations on both open-domain and enterprise benchmarks demonstrate that HarnessEvolve consistently outperforms state-of-the-art baselines across all five benchmarks.

\bibliographystyle{plainnat}
\bibliography{references}

@inproceedings{wei2022cot,
  author       = {Jason Wei and
                  Xuezhi Wang and
                  Dale Schuurmans and
                  Maarten Bosma and
                  Brian Ichter and
                  Fei Xia and
                  Ed H. Chi and
                  Quoc V. Le and
                  Denny Zhou},
  title        = {Chain-of-Thought Prompting Elicits Reasoning in Large Language Models},
  booktitle    = {NeurIPS},
  year         = {2022}
}

@inproceedings{yao2023react,
  author       = {Shunyu Yao and
                  Jeffrey Zhao and
                  Dian Yu and
                  Nan Du and
                  Izhak Shafran and
                  Karthik R. Narasimhan and
                  Yuan Cao},
  title        = {ReAct: Synergizing Reasoning and Acting in Language Models},
  booktitle    = {{ICLR}},
  publisher    = {OpenReview.net},
  year         = {2023}
}

@inproceedings{press2023selfask,
  author       = {Ofir Press and
                  Muru Zhang and
                  Sewon Min and
                  Ludwig Schmidt and
                  Noah A. Smith and
                  Mike Lewis},
  title        = {Measuring and Narrowing the Compositionality Gap in Language Models},
  booktitle    = {{EMNLP} (Findings)},
  series       = {Findings of {ACL}},
  volume       = {{EMNLP} 2023},
  pages        = {5687--5711},
  publisher    = {Association for Computational Linguistics},
  year         = {2023}
}

@inproceedings{yao2024tot,
  author       = {Shunyu Yao and
                  Dian Yu and
                  Jeffrey Zhao and
                  Izhak Shafran and
                  Tom Griffiths and
                  Yuan Cao and
                  Karthik Narasimhan},
  title        = {Tree of Thoughts: Deliberate Problem Solving with Large Language Models},
  booktitle    = {NeurIPS},
  year         = {2023}
}

@inproceedings{besta2024got,
  author       = {Maciej Besta and
                  Nils Blach and
                  Ales Kubicek and
                  Robert Gerstenberger and
                  Michal Podstawski and
                  Lukas Gianinazzi and
                  Joanna Gajda and
                  Tomasz Hoang and
                  Hubert Niewiadomski and
                  Piotr Nyczyk and
                  Torsten Hoefler},
  title        = {Graph of Thoughts: Solving Elaborate Problems with Large Language Models},
  booktitle    = {AAAI},
  year         = {2024}
}

@inproceedings{hong2024metagpt,
  author       = {Sirui Hong and
                  Mingchen Zhuge and
                  Jonathan Chen and
                  Xiawu Zheng and
                  Yuheng Cheng and
                  Jinlin Wang and
                  Ceyao Zhang and
                  Zili Wang and
                  Steven Ka Shing Yau and
                  Zijuan Lin and
                  Liyang Zhou and
                  Chenyu Ran and
                  Lingfeng Xiao and
                  Chenglin Wu and
                  J{\"{u}}rgen Schmidhuber},
  title        = {MetaGPT: Meta Programming for {A} Multi-Agent Collaborative Framework},
  booktitle    = {{ICLR}},
  publisher    = {OpenReview.net},
  year         = {2024}
}

@inproceedings{qian2024chatdev,
  author       = {Chen Qian and
                  Wei Liu and
                  Hongzhang Liu and
                  Nuo Chen and
                  Yufan Dang and
                  Jiahao Li and
                  Cheng Yang and
                  Weize Chen and
                  Yusheng Su and
                  Xin Cong and
                  Juyuan Xu and
                  Dahai Li and
                  Zhiyuan Liu and
                  Maosong Sun},
  title        = {ChatDev: Communicative Agents for Software Development},
  booktitle    = {{ACL} {(1)}},
  pages        = {15174--15186},
  publisher    = {Association for Computational Linguistics},
  year         = {2024}
}

@article{wu2023autogen,
  author       = {Qingyun Wu and
                  Gagan Bansal and
                  Jieyu Zhang and
                  Yiran Wu and
                  Shaokun Zhang and
                  Erkang Zhu and
                  Beibin Li and
                  Li Jiang and
                  Xiaoyun Zhang and
                  Chi Wang},
  title        = {AutoGen: Enabling Next-Gen {LLM} Applications via Multi-Agent Conversation
                  Framework},
  journal      = {CoRR},
  volume       = {abs/2308.08155},
  year         = {2023}
}

@inproceedings{schick2023toolformer,
  author       = {Timo Schick and
                  Jane Dwivedi{-}Yu and
                  Roberto Dess{\`{\i}} and
                  Roberta Raileanu and
                  Maria Lomeli and
                  Eric Hambro and
                  Luke Zettlemoyer and
                  Nicola Cancedda and
                  Thomas Scialom},
  title        = {Toolformer: Language Models Can Teach Themselves to Use Tools},
  booktitle    = {NeurIPS},
  year         = {2023}
}

@inproceedings{khattab2024dspy,
  author       = {Omar Khattab and
                  Arnav Singhvi and
                  Paridhi Maheshwari and
                  Zhiyuan Zhang and
                  Keshav Santhanam and
                  Sri Vardhamanan and
                  Saiful Haq and
                  Ashutosh Sharma and
                  Thomas T. Joshi and
                  Hanna Moazam and
                  Heather Miller and
                  Matei Zaharia and
                  Christopher Potts},
  title        = {DSPy: Compiling Declarative Language Model Calls into State-of-the-Art
                  Pipelines},
  booktitle    = {{ICLR}},
  publisher    = {OpenReview.net},
  year         = {2024}
}

@inproceedings{zhou2024webarena,
  author       = {Shuyan Zhou and
                  Frank F. Xu and
                  Hao Zhu and
                  Xuhui Zhou and
                  Robert Lo and
                  Abishek Sridhar and
                  Xianyi Cheng and
                  Tianyue Ou and
                  Yonatan Bisk and
                  Daniel Fried and
                  Uri Alon and
                  Graham Neubig},
  title        = {WebArena: {A} Realistic Web Environment for Building Autonomous Agents},
  booktitle    = {{ICLR}},
  publisher    = {OpenReview.net},
  year         = {2024}
}

@article{sumers2024cognitive,
  author       = {Theodore R. Sumers and
                  Shunyu Yao and
                  Karthik Narasimhan and
                  Thomas L. Griffiths},
  title        = {Cognitive Architectures for Language Agents},
  journal      = {Trans. Mach. Learn. Res.},
  volume       = {2024},
  year         = {2024}
}

@inproceedings{liu2024agentbench,
  author       = {Xiao Liu and
                  Hao Yu and
                  Hanchen Zhang and
                  Yifan Xu and
                  Xuanyu Lei and
                  Hanyu Lai and
                  Yu Gu and
                  Hangliang Ding and
                  Kaiwen Men and
                  Kejuan Yang and
                  Shudan Zhang and
                  Xiang Deng and
                  Aohan Zeng and
                  Zhengxiao Du and
                  Chenhui Zhang and
                  Sheng Shen and
                  Tianjun Zhang and
                  Yu Su and
                  Huan Sun and
                  Minlie Huang and
                  Yuxiao Dong and
                  Jie Tang},
  title        = {AgentBench: Evaluating LLMs as Agents},
  booktitle    = {{ICLR}},
  publisher    = {OpenReview.net},
  year         = {2024}
}

@inproceedings{ma2024agentboard,
  author       = {Chang Ma and
                  Junlei Zhang and
                  Zhihao Zhu and
                  Cheng Yang and
                  Yujiu Yang and
                  Yaohui Jin and
                  Zhenzhong Lan and
                  Lingpeng Kong and
                  Junxian He},
  title        = {AgentBoard: An Analytical Evaluation Board of Multi-turn {LLM} Agents},
  booktitle    = {NeurIPS},
  year         = {2024}
}

@inproceedings{zhao2024expel,
  author       = {Andrew Zhao and
                  Daniel Huang and
                  Quentin Xu and
                  Matthieu Lin and
                  Yong{-}Jin Liu and
                  Gao Huang},
  title        = {ExpeL: {LLM} Agents Are Experiential Learners},
  booktitle    = {{AAAI}},
  pages        = {19632--19642},
  publisher    = {{AAAI} Press},
  year         = {2024}
}

@inproceedings{shinn2023reflexion,
  author       = {Noah Shinn and
                  Federico Cassano and
                  Ashwin Gopinath and
                  Karthik Narasimhan and
                  Shunyu Yao},
  title        = {Reflexion: language agents with verbal reinforcement learning},
  booktitle    = {NeurIPS},
  year         = {2023}
}

@article{yang2023opro,
  author       = {Chengrun Yang and
                  Xuezhi Wang and
                  Yifeng Lu and
                  Hanxiao Liu and
                  Quoc V. Le and
                  Denny Zhou and
                  Xinyun Chen},
  title        = {Large Language Models as Optimizers},
  journal      = {CoRR},
  volume       = {abs/2309.03409},
  year         = {2023}
}

@inproceedings{pryzant2023protegi,
  author       = {Reid Pryzant and
                  Dan Iter and
                  Jerry Li and
                  Yin Tat Lee and
                  Chenguang Zhu and
                  Michael Zeng},
  title        = {Automatic Prompt Optimization with "Gradient Descent" and Beam Search},
  booktitle    = {{EMNLP}},
  pages        = {7957--7968},
  publisher    = {Association for Computational Linguistics},
  year         = {2023}
}

@article{wang2023voyager,
  author       = {Guanzhi Wang and
                  Yuqi Xie and
                  Yunfan Jiang and
                  Ajay Mandlekar and
                  Chaowei Xiao and
                  Yuke Zhu and
                  Linxi Fan and
                  Anima Anandkumar},
  title        = {Voyager: An Open-Ended Embodied Agent with Large Language Models},
  journal      = {Trans. Mach. Learn. Res.},
  volume       = {2024},
  year         = {2024}
}

@article{skillopt2026,
  author       = {Yifan Yang and
                  Ziyang Gong and
                  Weiquan Huang and
                  Qihao Yang and
                  Ziwei Zhou and
                  Zisu Huang and
                  Yan Li and
                  Xuemei Gao and
                  Qi Dai and
                  Bei Liu and
                  Kai Qiu and
                  Yuqing Yang and
                  Dongdong Chen and
                  Xue Yang and
                  Chong Luo},
  title        = {SkillOpt: Executive Strategy for Self-Evolving Agent Skills},
  journal      = {CoRR},
  volume       = {abs/2605.23904},
  year         = {2026}
}

@inproceedings{zhang2025aflow,
  author       = {Jiayi Zhang and
                  Jinyu Xiang and
                  Zhaoyang Yu and
                  Fengwei Teng and
                  Xionghui Chen and
                  Jiaqi Chen and
                  Mingchen Zhuge and
                  Xin Cheng and
                  Sirui Hong and
                  Jinlin Wang and
                  Bingnan Zheng and
                  Bang Liu and
                  Yuyu Luo and
                  Chenglin Wu},
  title        = {AFlow: Automating Agentic Workflow Generation},
  booktitle    = {{ICLR}},
  publisher    = {OpenReview.net},
  year         = {2025}
}

@article{liu2025sew,
  author       = {Siwei Liu and
                  Jinyuan Fang and
                  Han Zhou and
                  Yingxu Wang and
                  Zaiqiao Meng},
  title        = {{SEW:} Self-Evolving Agentic Workflows for Automated Code Generation},
  journal      = {CoRR},
  volume       = {abs/2505.18646},
  year         = {2025}
}

@inproceedings{xiong2006kmeans,
  author       = {Hui Xiong and
                  Junjie Wu and
                  Jian Chen},
  title        = {K-means clustering versus validation measures: a data distribution
                  perspective},
  booktitle    = {{KDD}},
  pages        = {779--784},
  publisher    = {{ACM}},
  year         = {2006}
}

@article{liu2019large,
  title={Large-Scale Long-Tailed Recognition in an Open World},
  author={Ziwei Liu and Zhongqi Miao and Xiaohang Zhan and Jiayun Wang and Boqing Gong and Stella X. Yu},
  journal={2019 IEEE/CVF Conference on Computer Vision and Pattern Recognition (CVPR)},
  pages={2532-2541},
  year={2019}
}

@inproceedings{feldman2020does,
author = {Feldman, Vitaly},
title = {Does learning require memorization? a short tale about a long tail},
booktitle = {Proceedings of the 52nd Annual ACM SIGACT Symposium on Theory of Computing},
publisher = {Association for Computing Machinery},
year = {2020}
}

@article{gepa2025,
  author       = {Lakshya A. Agrawal and
                  Shangyin Tan and
                  Dilara Soylu and
                  Noah Ziems and
                  Rishi Khare and
                  Krista Opsahl{-}Ong and
                  Arnav Singhvi and
                  Herumb Shandilya and
                  Michael J. Ryan and
                  Meng Jiang and
                  Christopher Potts and
                  Koushik Sen and
                  Alexandros G. Dimakis and
                  Ion Stoica and
                  Daniel Klein and
                  Matei Zaharia and
                  Omar Khattab},
  title        = {{GEPA:} Reflective Prompt Evolution Can Outperform Reinforcement Learning},
  journal      = {CoRR},
  volume       = {abs/2507.19457},
  year         = {2025}
}

@article{dunn2017searchqa,
  author       = {Matthew Dunn and
                  Levent Sagun and
                  Mike Higgins and
                  V. Ugur G{\"{u}}ney and
                  Volkan Cirik and
                  Kyunghyun Cho},
  title        = {SearchQA: {A} New Q{\&}A Dataset Augmented with Context from a
                  Search Engine},
  journal      = {CoRR},
  volume       = {abs/1704.05179},
  year         = {2017}
}

@inproceedings{zhang2026agentic,
  title={Agentic context engineering: Evolving contexts for self-improving language models},
  author={Zhang, Qizheng and Hu, Changran and Upasani, Shubhangi and Ma, Boyuan and Hong, Fenglu and Kamanuru, Vamsidhar and Rainton, Jay and Wu, Chen and Ji, Mengmeng and Li, Hanchen and others},
  booktitle={International Conference on Learning Representations},
  volume={2026},
  pages={86069--86100},
  year={2026}
}

@article{opsahl2026officeqa,
  title={Officeqa pro: An enterprise benchmark for end-to-end grounded reasoning},
  author={Opsahl-Ong, Krista and Singhvi, Arnav and Collins, Jasmine and Zhou, Ivan and Wang, Cindy and Baheti, Ashutosh and Oertell, Owen and Portes, Jacob and Havens, Sam and Elsen, Erich and others},
  journal={arXiv preprint arXiv:2603.08655},
  year={2026}
}

@article{zhang2026selfharness,
  title   = {Self-Harness: Harnesses That Improve Themselves},
  author  = {Zhang, Hangfan and Zhang, Shao and Li, Kangcong and Zhang, Chen and Chen, Yang and Zhang, Yiqun and Bai, Lei and Hu, Shuyue},
  journal = {arXiv preprint arXiv:2606.09498},
  year    = {2026}

}

@article{lin2026ahe,
  title   = {Agentic Harness Engineering: Observability-Driven Automatic Evolution of Coding-Agent Harnesses},
  author  = {Lin, Jiahang and Liu, Shichun and Pan, Chengjun and Lin, Lizhi and Dou, Shihan and Xi, Zhiheng and Huang, Xuanjing and Yan, Hang and Han, Zhenhua and Gui, Tao and Jiang, Yu-Gang},
  journal = {arXiv preprint arXiv:2604.25850},
  year    = {2026}

}

@article{liang2026skillnet,
  title   = {SkillNet: Create, Evaluate, and Connect AI Skills},
  author  = {Liang, Yuan and Zhong, Ruobin and Xu, Haoming and Jiang, Chen and Zhong, Yi and Fang, Runnan and Gu, Jia-Chen and Deng, Shumin and Yao, Yunzhi and Wang, Mengru and Qiao, Shuofei and Xu, Xin and Wu, Tongtong and Wang, Kun and Liu, Yang and Bi, Zhen and Lou, Jungang and Jiang, Yuchen Eleanor and Zhu, Hangcheng and Yu, Gang and Hong, Haiwen and Huang, Longtao and Xue, Hui and Wang, Chenxi and Wang, Yijun and others},
  journal = {arXiv preprint arXiv:2603.04448},
  year    = {2026}

}

@inproceedings{ma2024spreadsheetbench,
  author       = {Zeyao Ma and
                  Bohan Zhang and
                  Jing Zhang and
                  Jifan Yu and
                  Xiaokang Zhang and
                  Xiaohan Zhang and
                  Sijia Luo and
                  Xi Wang and
                  Jie Tang},
  title        = {SpreadsheetBench: Towards Challenging Real World Spreadsheet Manipulation},
  booktitle    = {NeurIPS},
  year         = {2024}
}

\newpage
\appendix

\section{Before-and-After Comparison of Harness Optimization}

\begin{figure}[ht]
    \centering
    \includegraphics[width=0.95\columnwidth]{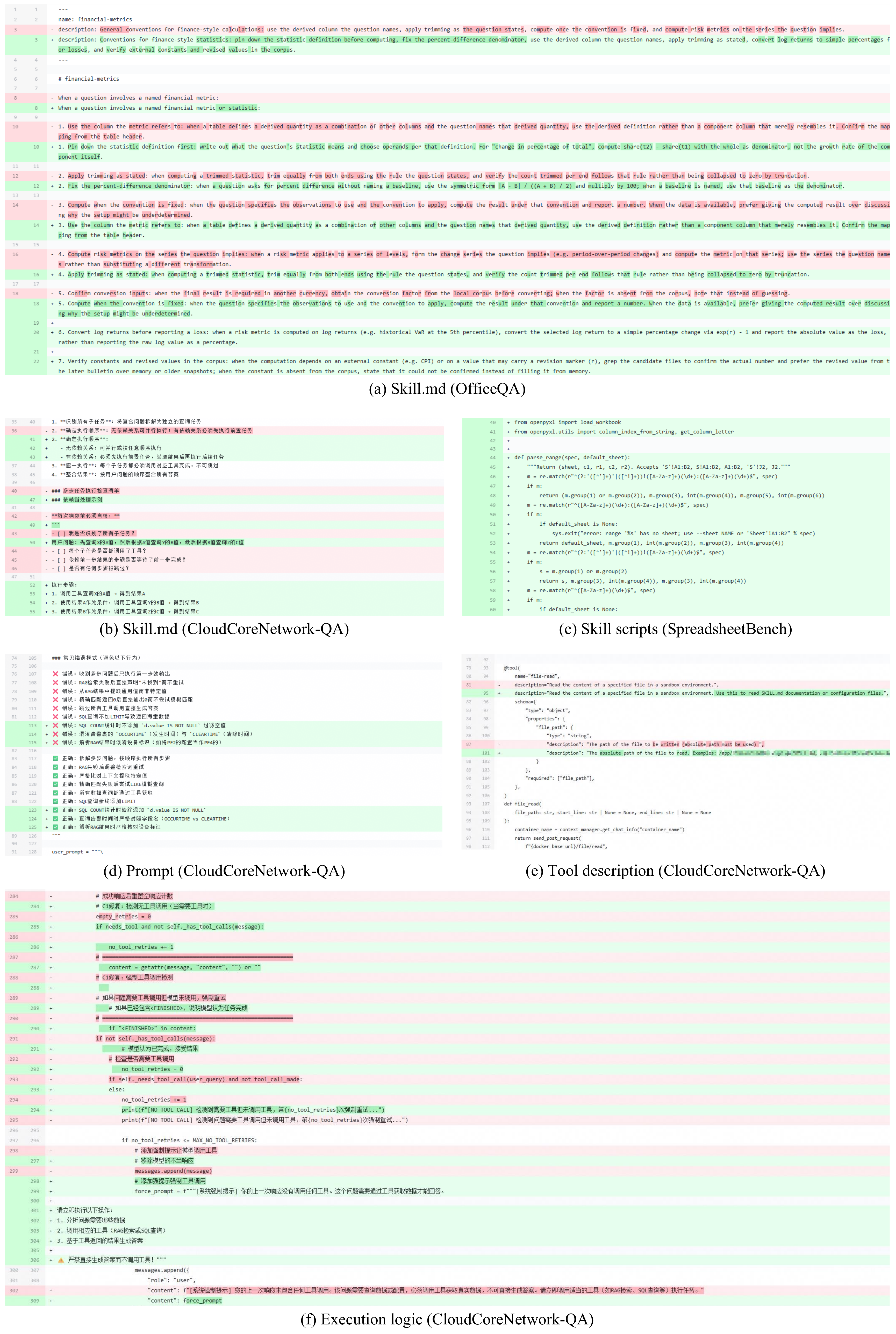}
    \caption{Before-and-after comparison of the harness optimization of HarnessEvolve across six cases: (a)~\texttt{skill.md} modification on OfficeQA, (b)~\texttt{skill.md} modification on CloudCoreNetwork-QA, (c)~script creation within the skill on SpreadsheetBench, (d)~system prompt modification on CloudCoreNetwork-QA, (e)~tool description modification on CloudCoreNetwork-QA, and (f)~execution logic modification on CloudCoreNetwork-QA. HarnessEvolve modifies the entire execution agent---including \texttt{skill.md} files, scripts, prompts, tool descriptions, and execution-logic code---rather than optimizing a single component in isolation.}
    \label{fig:optimization}
\end{figure}

Figure~\ref{fig:optimization} compares the harness before and after the optimization of HarnessEvolve across six cases: (a)~\texttt{skill.md} modification on OfficeQA, (b)~\texttt{skill.md} modification on CloudCoreNetwork-QA, (c)~script creation within the skill on SpreadsheetBench, (d)~system prompt modification on CloudCoreNetwork-QA, (e)~tool description modification on CloudCoreNetwork-QA, and (f)~execution logic modification on CloudCoreNetwork-QA.

\section{Reference-Guided Error Diagnosis and Clustering}

Algorithm~\ref{alg:unified_pipeline_clean} details the error diagnosis and clustering procedure. For each failed instance, the optimization agent retrieves the corresponding reference trajectory from the global cache $\mathcal{C}$ and performs trajectory comparison to identify the first action divergence point as the root cause. When no reference trajectory is available, the agent falls back to isolated trajectory analysis. The collected error signals are then grouped via error clustering to preserve rare failure patterns.

\begin{algorithm}[ht]
    \caption{Reference-Guided Error Diagnosis and Clustering}
    \label{alg:unified_pipeline_clean}
    \begin{algorithmic}[1]
        \Require Failed instances $F_j = \{ (q_i, \tau^-_i) \}$, global cache $\mathcal{C}$ (containing verified reference trajectories $\tau^+_i$ for queries that yielded a genuine execution path).
        \Ensure Error groups $\mathcal{P} = \{C_1, \dots, C_K\}$, where each $C_k = (\bar{s}_k, \bar{m}_k, \bar{e}_k, \bar{h}_k)$.

        \State Initialize error batch $\mathcal{M} \leftarrow \emptyset$.

        \For{$i = 1$ \textbf{to} $|F_j|$}
            \State $\tau^+_i \leftarrow \mathcal{C}[q_i]$. \Comment{look up pre-computed reference}

            \If{$\tau^+_i = \emptyset$}
                \State $\mathcal{F}_i \leftarrow \mathrm{AnalyzeTrajectory}(\tau^-_i, q_i)$. \Comment{without reference}
            \Else
                \State $\mathcal{F}_i \leftarrow \mathrm{AnalyzeTrajectory}(\tau^-_i, \tau^+_i, q_i)$. \Comment{compare for fine-grained root cause}
            \EndIf
            \State $\mathcal{M} \leftarrow \mathcal{M} \cup \{\mathcal{F}_i\}$.
        \EndFor

        \State $\mathcal{P} \leftarrow \mathrm{ErrorCluster}(\mathcal{M})$.

        \State \Return $\mathcal{P}$
    \end{algorithmic}
\end{algorithm}

\section{HarnessEvolve Self-Evolution Procedure}

Algorithm~\ref{alg:evolution_loop_clean} presents the complete self-evolution loop. Each epoch partitions the training set into mini-batches and iterates through them. For each batch, the optimization agent generates candidate updates, the gate agent filters them, and the performance gate decides acceptance based on improvement margin $\delta$ and replay degradation $\epsilon$. Accepted snapshots are added to the pool $\mathbb{V}$. At epoch end, the best-performing snapshot on the validation set is selected as $\mathcal{A}^*$. The loop terminates early when either batch-level patience $P_{\mathrm{batch}}$ or epoch-level patience $P_{\mathrm{ep}}$ is exhausted.

\begin{algorithm}[ht]
    \caption{HarnessEvolve Self-Evolution Procedure}
    \label{alg:evolution_loop_clean}
    \begin{algorithmic}[1]
        \Require Task corpora $\mathcal{T}_\mathrm{train}, \mathcal{T}_\mathrm{val}, \mathcal{T}_\mathrm{test}$, agents $(\mathcal{A}_{\mathrm{exec}}, \mathcal{A}_{\mathrm{eval}}, \mathcal{A}_{\mathrm{opt}}, \mathcal{A}_{\mathrm{gate}})$, global cache $\mathcal{C}$ (containing verified reference trajectories $\tau^+_i$ for queries that yielded a genuine execution path), batch patience $P_{\mathrm{batch}}$, epoch patience $P_{\mathrm{ep}}$, step margin $\delta \geq 0$, degradation tolerance $\epsilon \geq 0$, replay buffer size $R$, filter bounds $(\eta_{\mathrm{leak}}, \eta_{\mathrm{blo}})$, epoch limit $E$, batch size $b$, revision limit $T_{\mathrm{rev}}$.
        \Ensure Optimized agent $\mathcal{A}^*$, snapshot pool $\mathbb{V}$.

        \State Initialize $\mathcal{A}_{\mathrm{current}} \leftarrow \mathcal{A}_{\mathrm{exec}}$, $\mathcal{A}^* \leftarrow \mathcal{A}_{\mathrm{exec}}$, snapshot pool $\mathbb{V} \leftarrow \{\mathcal{A}_{\mathrm{exec}}\}$, index $n \leftarrow 0$, rejection count $c \leftarrow 0$, best validation accuracy $a_{\mathrm{best}} \leftarrow \mathrm{Acc}(\mathcal{A}_{\mathrm{exec}}, \mathcal{T}_\mathrm{val})$, epoch non-improvement count $c_{\mathrm{ep}} \leftarrow 0$.

        \For{$e = 1$ \textbf{to} $E$}
            \State $\mathcal{B} \leftarrow \mathrm{PermuteAndPartition}(\mathcal{T}_\mathrm{train}, b)$

            \For{each batch $B_j \in \mathcal{B}$}
                \State Run $\mathcal{A}_{\mathrm{current}}$ on $B_j$ to produce trajectories $\{\tau_i\}$
                \State $F_j \leftarrow \mathcal{A}_{\mathrm{eval}}(\{\tau_i\}, B_j)$ \Comment{identify failed instances}
                \State $\mathcal{P} \leftarrow \mathrm{DiagnoseAndCluster}(F_j, \mathcal{C})$ \Comment{Algorithm~\ref{alg:unified_pipeline_clean}}
                \State $\Delta_{\mathrm{harness}} \leftarrow \mathcal{A}_{\mathrm{opt}}(\mathcal{P}, \mathcal{A}_{\mathrm{current}})$
                \State $\Delta_{\mathrm{verified}}, \rho \leftarrow \mathcal{A}_{\mathrm{gate}}(\Delta_{\mathrm{harness}}; \eta_{\mathrm{leak}}, \eta_{\mathrm{blo}})$
                \For{$r = 1$ \textbf{to} $T_{\mathrm{rev}}$}
                    \If{$\Delta_{\mathrm{verified}} \neq \emptyset$}
                        \State \textbf{break}
                    \EndIf
                    \State $\Delta_{\mathrm{harness}} \leftarrow \mathcal{A}_{\mathrm{opt}}(\mathcal{P}, \mathcal{A}_{\mathrm{current}}, \rho)$ \Comment{revise with rejection feedback}
                    \State $\Delta_{\mathrm{verified}}, \rho \leftarrow \mathcal{A}_{\mathrm{gate}}(\Delta_{\mathrm{harness}}; \eta_{\mathrm{leak}}, \eta_{\mathrm{blo}})$
                \EndFor
                \If{$\Delta_{\mathrm{verified}} = \emptyset$}
                    \State \textbf{continue}
                \EndIf

                \State $\mathcal{A}_{\mathrm{candidate}} \leftarrow \mathrm{Build}(\mathcal{A}_{\mathrm{current}}, \Delta_{\mathrm{verified}})$
                \State $\Delta_{\mathrm{Acc}} \leftarrow \mathrm{Acc}(\mathcal{A}_{\mathrm{candidate}}, B_j) - \mathrm{Acc}(\mathcal{A}_{\mathrm{current}}, B_j)$ 
                \State $\Delta_{\mathrm{deg}} \leftarrow \max_{l \in [\max(1,\, j-R),\, j-1]} [\mathrm{Acc}(\mathcal{A}_{\mathrm{current}}, B_l) - \mathrm{Acc}(\mathcal{A}_{\mathrm{candidate}}, B_l)]$ 
                \If{$\Delta_{\mathrm{Acc}} \ge \delta$ \textbf{and} $\Delta_{\mathrm{deg}} \le \epsilon$} \Comment{accept}
                    \State $n \leftarrow n + 1$;
                    \State $\mathcal{A}^{(n)} \leftarrow \mathcal{A}_{\mathrm{candidate}}$; $\mathbb{V} \leftarrow \mathbb{V} \cup \{\mathcal{A}^{(n)}\}$; $\mathcal{A}_{\mathrm{current}} \leftarrow \mathcal{A}^{(n)}$; $c \leftarrow 0$
                \Else \Comment{reject}
                    \State $c \leftarrow c + 1$
                \EndIf

                \If{$c \ge P_{\mathrm{batch}}$}
                    \State \textbf{break}
                \EndIf
            \EndFor

            \State $\mathcal{A}_{\mathrm{current}} \leftarrow \arg\max_{\mathcal{A} \in \mathbb{V}} \mathrm{Acc}(\mathcal{A}, \mathcal{T}_\mathrm{val})$; $\mathcal{A}^* \leftarrow \mathcal{A}_{\mathrm{current}}$
            \If{$\mathrm{Acc}(\mathcal{A}^*, \mathcal{T}_\mathrm{val}) > a_{\mathrm{best}}$}
                \State $a_{\mathrm{best}} \leftarrow \mathrm{Acc}(\mathcal{A}^*, \mathcal{T}_\mathrm{val})$; $c_{\mathrm{ep}} \leftarrow 0$
            \Else
                \State $c_{\mathrm{ep}} \leftarrow c_{\mathrm{ep}} + 1$
            \EndIf
            \If{$c_{\mathrm{ep}} \ge P_{\mathrm{ep}}$}
                \State \textbf{break}
            \EndIf
        \EndFor

        \State \Return $\mathcal{A}^*$ \Comment{Report $\mathrm{Acc}(\mathcal{A}^*, \mathcal{T}_\mathrm{test})$}
    \end{algorithmic}
\end{algorithm}

\end{document}